\documentclass[sigconf]{acmart}

\usepackage{booktabs} 
\usepackage{url}
\usepackage{subfig}

\begin{document}
\title[Pairing Character Classes via a Deep-Learning Surrogate Model]{Pairing Character Classes in a Deathmatch Shooter Game via a Deep-Learning Surrogate Model}
\subtitle{}

\author{Daniel Karavolos}
\email{daniel.karavolos@um.edu.mt}
\affiliation{%
  \institution{Institute of Digital Games\\ University of Malta}
  \country{Malta}
}
\author{Antonios Liapis}
\email{antonios.liapis@um.edu.mt}
\affiliation{%
  \institution{Institute of Digital Games\\ University of Malta}
  \country{Malta}
}
\author{Georgios N. Yannakakis}
\email{georgios.yannakakis@um.edu.mt}
\affiliation{%
  \institution{Institute of Digital Games\\ University of Malta}
  \country{Malta}
}

\copyrightyear{2018}
\acmYear{2018}
\setcopyright{acmlicensed}
\acmConference[FDG'18]{FDG 2018}{August 7--10, 2018}{Malm\"{o}, Sweden}
\acmBooktitle{Foundations of Digital Games 2018 (FDG'18), August 7--10, 2018, Malm\"{o}, Sweden}
\acmPrice{15.00}
\acmDOI{10.1145/3235765.3235816}
\acmISBN{978-1-4503-6571-0/18/08}


\begin{abstract}
This paper introduces a surrogate model of gameplay that learns the mapping between different game facets, and applies it to a generative system which designs new content in one of these facets. Focusing on the shooter game genre, the paper explores how deep learning can help build a model which combines the game level structure and the game's character class parameters as input and the gameplay outcomes as output. The model is trained on a large corpus of game data from simulations with artificial agents in random sets of levels and class parameters. The model is then used to generate classes for specific levels and for a desired game outcome, such as balanced matches of short duration. Findings in this paper show that the system can be expressive and can generate classes for both computer generated and human authored levels.
\end{abstract}

\maketitle

\section{Introduction}\label{sec:introduction}

When two game designers talk about a new game concept, they do not have to write down all the rules or play a prototype of that game in order to understand each other. They can make an estimate of the resulting gameplay based on only a few pieces of information by extrapolating from past experiences, or by using examples of games that they both know well. In a level for a shooter game, for example, it is easy for a designer or a veteran player to spot choke points or an unfair distribution of powerups (e.g. based on distance to the respective team bases) by looking at the top-down map.

Based on their own expert knowledge, several researchers have identified key game patterns \cite{bjork2004patterns} or game level patterns \cite{hullet2010fpspatterns}. Moreover, many academic papers have formulated properties of game levels \cite{liapis2013evaluations} or rulesets \cite{browne2010boardgames} in a quantitative fashion, often to use them as an objective to optimize towards \cite{togelius2011searchbased}. However, it is arguably infeasible to formalize and accurately compute all possible metrics of level quality. Not only are some metrics difficult to capture, but how they impact game quality can be influenced by the nuances of the specific game's ruleset. Moreover, each game level can be evaluated in any number of dimensions, e.g. in terms of visuals such as symmetry and in terms of game-specific patterns such as exploration or balance \cite{preuss2014gooddiverse}. It is not straightforward how to combine these metrics in a way that the computer can understand or produce content that optimizes them. For instance, aggregating all features (with equal weights) into one fitness that incorporates multiple level metrics \cite{liapis2013evaluations} is a necessary over-simplification that a human designer would not need to make.

Beyond patterns in a single facet of games (such as level patterns for level design), a game constitutes a multi-faceted product and experience \cite{liapis2014gamecreativity}. Games are more than the sum of their parts (levels, sounds, graphics, rules): the background soundtrack influences the mood of a horror game as much as a ghost story that is told through the game or the dark winding corridors of the mansion the player must navigate \cite{lopes2016framing}. Human game designers understand the interrelations \emph{between} facets and how changes in one may influence all others. However, these relationships and causal chains are more difficult to capture computationally.

Driven by a desire to instill a more human-like creative agency to computational game designers, we identify an important stepping stone in the automated assessment of game content less as parts and more as a whole \cite{bidarra2015orchestration}. Being able to identify interrelations between, for instance, game rules and level structures without needing to produce and test the outcome can allow such a computational designer to reason in a more similar fashion to a human designer. In turn, this opens the possibility of a more intuitive dialog \cite{novick1997mixedinit} between a human and a computational designer where the artificial intelligence can suggest or undertake changes in one facet based on human input in the same or a different facet.

Towards realizing such a computationally creative game designer, this paper presents a system which can learn the relationships between level design, game design parameters, and the outcomes of gameplay. Similar to a human designer's perception, the system uses the visual description of a game level as an image and learns to see the important level patterns that can affect the balance or quality of a playthrough. This paper focuses on first person shooter games, as they are straightforward to both understand and simulate and have been the focus of game design \cite{hullet2010fpspatterns} and generation \cite{cardamone2011fps}. While past research projects have focused on generating weapons \cite{gravina2015weapons}, the game design space explored in this paper also includes parameters of the player's agent. \emph{Character classes} are common in shooter games: each competing player chooses a class, which may have a different survivability and speed as well as a signature weapon which fits its role (e.g. a scout or a medic class). Classes and their weapons are the most sensitive aspects of a shooter game's ruleset, and are often tweaked even after the game is launched.

In this constrained yet expressive design space, the mapping between the different facets is learned through deep neural networks, using convolution for recognizing elements of the level's top down map (provided as an image) and with class parameters as additional input in order to predict the game's core gameplay outcomes: the winning player and the duration of the game. Using an extensive corpus of simulated playthroughs with a broad range of level structures and class parameters, the convolutional neural network is able to identify patterns better than other machine learning baselines. As a proactive designer, the system exploits its past experiences (i.e. the learned model) to create classes for each player in a specific two-player level. These classes are tuned towards a specific match duration and towards balance between players, using the deep network as a \emph{surrogate model} \cite{jin2011surrogate} of gameplay instead of computationally expensive simulations. The paper shows how the system can generate classes that satisfy the intended design goals, for game levels that are human-authored or computer-generated. The software therefore demonstrates that it can be used intuitively as a creative companion for a human level designer or paired with another computational designer (a level designer, in this case). Both of these directions will be explored in future work.

\section{Related Work}

Increased attention in the research and practice of procedural content generation has seen a vast increase in the range and quality of generators. While historically generation in games focused on automating the process of level design, in order to present the player with new environments in games such as \emph{Rogue} (Toy and Wichman 1980), \emph{ELITE} (Acornsoft 1984) and \emph{Diablo} (Blizzard 1996), the types of game content created today are much more varied. Games are built on a variety of content, the creation of which requires different but complementary types of creative input: \cite{liapis2014gamecreativity} identified six facets of games in terms of visuals, audio, narrative, levels, rules, and gameplay. Content belonging to each of these facets has been generated in several capacities (academic or commercial) and using a variety of algorithmic approaches \cite{shaker2016procedural}. While most of these generators focus on one facet (such as a level generator, a sound effect generator etc.), a challenging but important direction for content generation lies in the orchestration of these facets \cite{bidarra2015orchestration} so that different generators can combine their outputs into a cohesive, harmonious whole. An early embryo of such multi-faceted generation evolved rules and levels \cite{cook2011multi}, evaluating the combined outcome using agent-based simulations. Similar simulation-based evaluations abound in the PCG literature \cite{togelius2011searchbased}; their main limitation is that they are very computationally expensive. For more complex games and/or more elaborate AI controllers, having to complete one or more simulations to assess a single individual in an evolving population becomes  unrealistic.

This paper explores how machine learning can be used for procedural content generation as a surrogate model, indirectly influencing the fitness function of a search-based PCG algorithm \cite{togelius2011searchbased}. So far, machine learned models are primarily used to directly manipulate game content \cite{summerville2017procedural}. For instance, neural networks have mostly been used to learn level patterns which are then applied directly to the level. For example, a recurrent neural network that predicts sequences of tiles is used to create levels for \emph{Super Mario Bros.} (Nintendo 1985) in \cite{summerville2016super}; a convolutional neural network (CNN) is used to place resources on a pre-made \emph{Starcraft II} (Blizzard 2010) map \cite{lee2016predicting}. Other work has used autoencoders to learn patterns in \emph{Super Mario Bros.} levels, taking advantage of the encoding-decoding sequence to repair broken segments \cite{jain2016autoencoders}. Finally, CNNs have been used to predict various characteristics (difficulty, enjoyment and aesthetics) of \emph{Super Mario Bros.} levels based on player annotations \cite{guzdial2016deep}, but these networks were not used for content generation. 

While machine learning has a long history in procedural generation \cite{summerville2017procedural}, the proposed framework uses its learned model indirectly (as a surrogate model to guide evolution) rather than directly. More importantly, it follows earlier research \cite{karavolos2017patterns} in merging game rules (in the form of class parameters) and level properties as inputs, in order to learn how their interrelations affect gameplay outcomes. The model therefore combines three different facets of games: rules, levels and gameplay as discussed in \cite{liapis2014gamecreativity}. The framework is the first step towards game facet orchestration where all facets of games are considered as a whole rather than e.g. considering only the structural parts of the level \cite{liapis2013evaluations} or the properties of weapons \cite{gravina2015weapons} in a vacuum. Compared to earlier work \cite{karavolos2017patterns}, the model introduced in this paper uses a much broader corpus of structurally and ludically complex levels and a diverse set of classes as its training set. It is also the first instance that uses such a multi-faceted computational model as a surrogate in generating aspects of a game's ruleset for specific levels and intended gameplay outcomes, while in \cite{karavolos2018surrogate} a similar surrogate model is used to drive the evolutionary adaptation of levels in order to balance a specific set of character classes.
	
The application domain in this paper is a shooter game; due to the lack of high quality open data for contemporary shooter games, the corpus is created via simulations of artificial agents in a simple two-player game. Shooter games have been the focus of both game design and game generation. Specifically, several attempts at game balancing have focused on changing the parameters of weapons \cite{gravina2015weapons,gravina2016constrainedsurprise} but also levels \cite{cardamone2011fps,lanzi2014maps}. Similar to this line of research, we aim to generate new weapons but also non-weapon attributes such as players' hit points. Moreover, rather than using expensive simulations, the associations learned by the model can be used to quickly explore a vast design space while being flexible in accounting for different level formats, including human-authored levels.

\section{Game Framework}\label{sec:framework}
As discussed in the introduction, this paper chooses first person shooter (FPS) games as a case study for mapping level structure and game parameters with gameplay outcomes, and for exploiting that mapping to create classes appropriate for a level. The FPS game framework specifically targets a one-versus-one death match: the goal of a death match is for one player\footnote{Using arcade game terms, we identify ``player'' in this paper as the virtual avatar, controlled by a human, which acts within the game. Evidently, no human players are hurt in these shooter games.} to kill the other player's avatar more times. In this study, the game ends after a total of 20 kills: the winner is the player with the most kills scored. 

The game used for these experiments is implemented in Unity 3D, a commercial game engine, and based on an existing toolkit \cite{opsive}. The game is played between two competing players that start in opposite sides of a game level (the characteristics of which are described below). Each player belongs to a character class: character classes are common in shooter games such as \emph{Team Fortress 2} (Valve 2007) and have different gameplay styles and strategies, as well as a different signature weapon. While game parameters for character classes and their weapons are described below, an important parameter is the hit points (HP): if any player drops to 0 HP, they are killed and the match comes closer to ending.

\subsection{Level format}

\begin{figure}[t]
	\centering					
	\subfloat[3D shooter level\label{subfig1:top_down}]{%
		\includegraphics[height=0.28\textwidth]{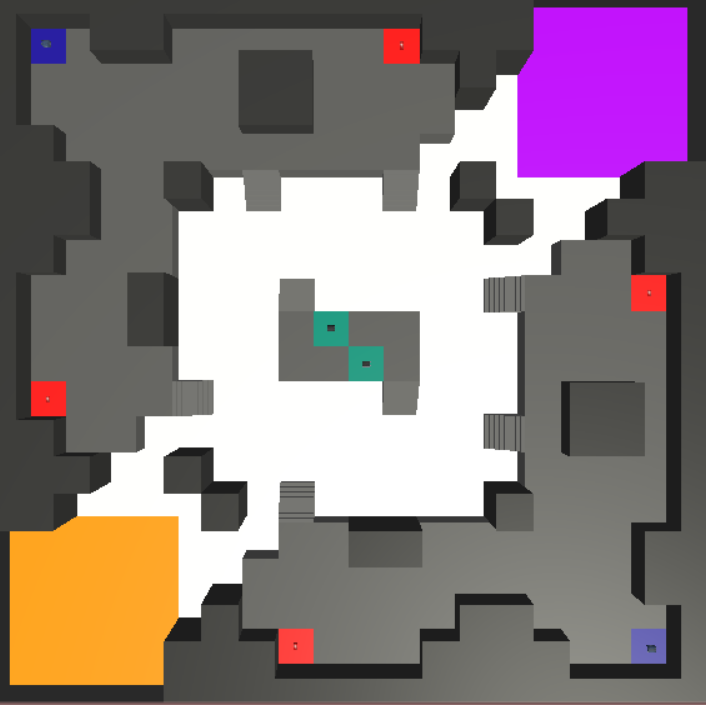}
	}
	\quad
	\subfloat[CNN inputs\label{subfig2:cnn_view}]{%
		\includegraphics[trim={23cm 0 1cm 0},clip,height=0.28\textwidth]{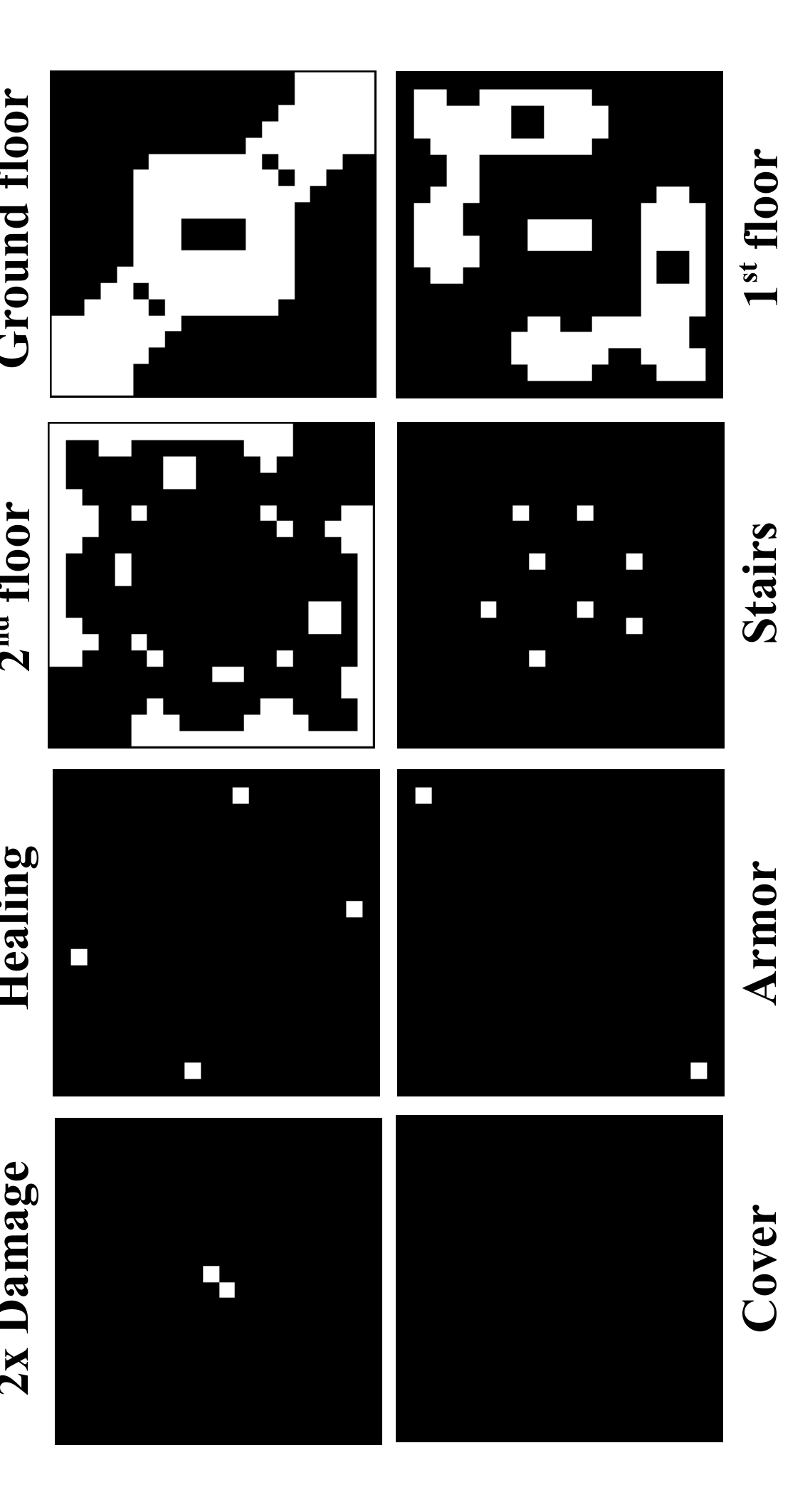}
	}
	
	\caption{A view of the in-game 3D level and its transformation into CNN inputs. Orange and purple areas are the bases of player 1 and 2 respectively. Red tiles are healing locations, blue and turquoise tiles are armor and double damage powerups respectively.
    }
	\label{fig:screenshot}
\end{figure}

Matches in this paper take place in two-floor levels. For the purposes of training our model of gameplay, we input the level as a visual representation of its top down map. Finally, to produce the necessary corpus of gameplay data, we use a straightforward level generator to create a large variety of shooter levels which are ensured to be playable. All of these aspects are detailed below.

\subsubsection{Map properties:}

Shooter game levels within this research are based on a grid of $20\times20$ tiles. Each tile has an elevation and can also feature a special entity. Tiles of different elevations form floors which players can traverse. Tiles at the ground elevation are the most common: all players start on the ground floor. Tiles on the first floor can be accessed from the ground floor via stairs, while players can jump down from the first floor to any adjacent ground floor tile. The second floor is not accessible via stairs, and acts as an impenetrable wall that players can not go through (or shoot through). Stairs can only be placed on the ground floor and allow access to a specific first floor tile which is adjacent to the stairs. Other special entities are powerups for the player passing through that tile: (1) \emph{double damage} for a temporary increase in damage output of the player's weapon, (2) \emph{healing} for restoring 100 of the player's HP, and (3) \emph{armor} which provides an additional 50 HP which are depleted first before the player needs to lose any of their actual HP. Once these powerups are taken, they respawn after a time delay which depends on the powerup. Finally, the players start the game from different corners of the map, and these $5\times5$ tile ``bases'' are color-coded (see Fig.~\ref{fig:screenshot}),  devoid of any other special entities, and always on the ground floor.

\subsubsection{Network input:}

To use the level's structure as input, the top-down map is converted into a form of one-hot encoding to make it easier for the network to process. From the visual top-down map (see Fig.~\ref{fig:levels}) we extract 8 binary channels of $20\times20$ pixels: three channels depict the three elevations of each tile. The stairs and each powerup type have their own channels which denote their positions (as 1). The 8$^{th}$ channel depicts cover positions, which were omitted (treated as 0) in this study.

\subsubsection{Level generation:}
To produce the corpus of simulated play-throughs, a map generator was used following a hierarchical digger agent approach \cite{shaker2016constructive}. First, generation operates on a low-resolution sketch consisting of $4\times4$ cells with the players' bases at the corners. Two agents randomly dig paths from the base of player 1 to the base of player 2, on each side of the bases' diagonal. Each cell of this sketch is then translated into $5 \times 5$ tiles (initially walls) for the final representation: a random digger digs paths in each cell according to the connections in the sketch. The algorithm computes the possible positions for stairs between ground and first floor, placing them with a 20\% probability. The walkable areas of the first floor are shaped by stochastic cellular automata \cite{johnson2010caves} that build the second floor (walls) on top of the first floor using patterns of the first floor. Finally, each of the original $4\times4$ sketch cells (except the bases) has a 33\% probability of receiving a powerup, which is placed randomly at one of its ground or first floor tiles. Note that this constructive approach ensures the connectivity of all walkable areas. 

\subsection{Game parameters}
As noted earlier, the goal of this paper is to generate an appropriate character class for each player based on the level and an intended game outcome. Each class has 2 parameters, i.e. hit points (HP) and movement speed, and its signature weapon has 6 parameters: damage (per shot), accuracy (i.e. probability and extent of dispersion), rate of fire (R.o.F), clip size, number of bullets per shot and weapon range. Players have infinite ammo, but will delay shooting in order to reload once their clip is depleted. Reload time is the same for each weapon regardless of other parameters. Each player always uses the weapon of their class; no extra weapons are available.

To instantiate new classes when producing the corpus of simulated playthroughs, each class parameter was normalized to a predefined, intuitive range (using the \emph{Team Fortress 2} game as a reference point) and a new value was generated randomly within that range. The exception was weapon range, where the possible values were long, medium or short (no intermediate values were allowed). For the sake of all experiments, game parameters used as input by the computational model and the class generator were min-max normalized to $[0,1]$ based on that predefined range.

\subsection{Simulated gameplay}
To produce a corpus which maps class parameters and level structure to gameplay outcomes, it is obviously necessary to simulate the match and find its outcomes. To do this, AI agents were used to play a match until 20 kills were scored or until a time limit was reached. The agents' behavior is controlled by behavior trees, which were adapted from those in the toolkit \cite{opsive}. The agents can detect the opponent through both sight and hearing. Map knowledge is simulated by allowing the agents to perceive all pickups within a wide radius around them. In descending order of priority, agents (a) search for healing items if their health is low, (b) pick up nearby powerups, (c) attack the opponent if one is in sight, and (d) search for far away powerups. While their underlying logic is straightforward, the agents' in-game behavior was deemed complex enough to act as an approximation to human play.

\section{Modeling Gameplay}\label{sec:modeling}
This section discusses the training data collected from artificial gameplays and analyzes how different computational models learn this data. The most successful model is used for generating character classes in Section \ref{sec:generating}.

\subsection{Training data}
In order to train any machine learning algorithm, it is vital to establish a corpus of rich and expressive data points which capture the relationships between level structures, weapon parameters and gameplay data. For this paper, our training data consists of $2{\cdot}10^5$ matches, using artificial agents to simulate human gameplay patterns. The data was collected as follows: a new map and a new pair of character classes was generated using the generative processes described in Section \ref{sec:framework}, and two matches were simulated with this setup, the first match with one class used by player 1 (orange corner of the map in Fig.~\ref{fig:screenshot}) and the second match with the same class used by player 2 (purple corner of the map in Fig.~\ref{fig:screenshot}). In this fashion, $10^5$ unique maps and character class pairs were tested. Each simulated match ended after a total of 20 kills, at which point the duration of the match and the kill ratio of each team was logged. Some matches could not be completed within a time limit, as agents could not score enough kills, and were omitted from the training data (6\% of the total data points).

\begin{figure}[!t]
	\centering
	\subfloat[Score\label{subfig2:score_hist}]{%
		\includegraphics[width=0.23\textwidth]{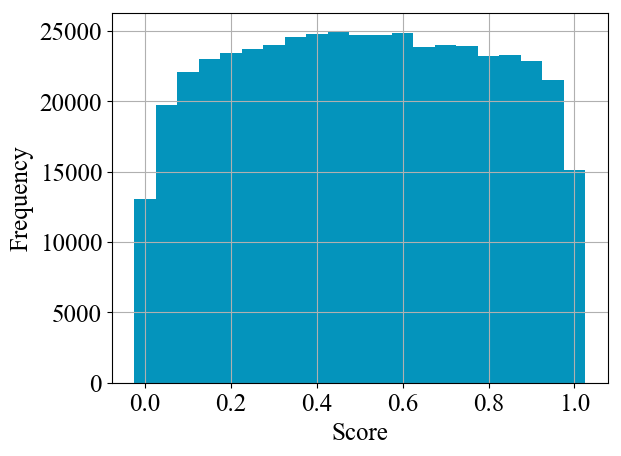}
	}
    	\subfloat[Game Duration\label{subfig1:time_hist}]{%
		\includegraphics[width=0.23\textwidth]{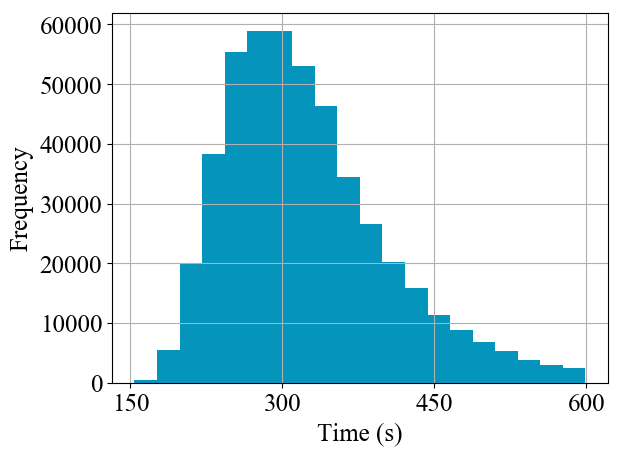}
	}
\caption{Distribution of gameplay outcomes in the corpus. }
\label{fig:training_data}	
\end{figure}

How these gameplay metrics are distributed is shown in Figure \ref{fig:training_data}. Game duration lies between 150 and 600 seconds, broadly following a skewed bell curve. It is evident from Fig. \ref{subfig1:time_hist} that most matches lasted approximately 300 seconds (mean of 323, standard deviation of 80). For the purpose of simplifying the target function for the model, game duration was normalized into the range of $[0,1]$ based on min-max normalization (from 150 to 600 sec).
It is evident from Fig.~\ref{subfig2:score_hist} that the score of player 1 is spread over the entire spectrum, from extreme advantage over player 2 (scores near 1.0) to extreme disadvantage (scores near $0.0$) and a balanced match (score of $0.5$). Unlike \cite{karavolos2017patterns}, the score does not follow a normal distribution centered around $0.5$; the distribution is rather uniform, with occurrences decreasing towards the edge cases. This points to a rich dataset with positive and negative examples of a balanced match. 

\subsection{Convolutional Network Architecture}
Based on \cite{karavolos2017patterns,karavolos2018surrogate} and additional preliminary experiments with different network architectures, learning rates and activation functions, the best performing CNN architecture used for reported experiments has two separate branches: one for level structure and one for class parameters. The character class branch consists of a single fully-connected layer of 8 nodes that transforms the 16 character class parameters (8 per player) into 8 features. The map input branch consists of two blocks of one convolution layer and one max-pooling layer each. The first block has 16 filters, the second block has 32 filters. Each convolutional filter is of size $5{\times}5$, and zero-padding is used to keep the output size same as the input size. The output of the second block is flattened into a vector of 800 level features. The output of both branches is concatenated and processed by one fully-connected decision layer of 128 nodes and two outputs. The two outputs are (a) kill ratio of player 1 (i.e. score) and (b) normalized game duration. All network nodes have exponential linear units (ELU) \cite{clevert2016fast} as their activation function.

\subsection{Model Validation}

With a training data set of almost $2{\cdot}10^5$ data points, there are many options for modeling the data via machine learning. As in \cite{karavolos2017patterns}, our core hypothesis is that a convolutional neural network (CNN) will be better at processing the level's layout as an image and combined with the class parameters as additional input in the final layers could lead to increased training accuracy. As baselines for this hypothesis, several multi-layer perceptron models (MLP) are tested, as well as a perceptron and linear regression (LR). For the sake of brevity the performance of the best MLP (with 16 neurons with ELU activations) is reported, although other MLP architectures we tested consisted of one or two layers with 2 to 1024 nodes each, and various activation functions. Unlike \cite{karavolos2017patterns}, the learning task is one of regression: the goal is to build a model which can output both a predicted game duration and the balance of a game (as the advantage in terms of score for player 1), and these two gameplay metrics are the outputs of the model.

\begin{table}
\centering
\caption{Average Mean Absolute Error (MAE) and the $R^{2}$ values for duration ($t$) and score ($s$) prediction based on a random validation set of 10\% of the corpus.}
\label{tab:training_results}
\begin{tabular}{l|cc|cc}
Network					& $MAE_t$ & $MAE_s$	&$R^{2}_t$ & $R^{2}_s$	\\
\hline
CNN        				& 0.082		& 0.068	& 0.60		& 0.90		\\ 
MLP    					& 0.085		& 0.069	& 0.57		& 0.89		\\ 
Perceptron 				& 0.93		& 0.088	& 0.50		& 0.84		\\ 
LR						& 0.095		& 0.095	& 0.47		& 0.82		
\end{tabular}
\end{table}

In the results reported in this section, all networks were trained for 100 epochs, using early stopping (with a patience of 5 epochs) to prevent overfitting. We can evaluate regression models by the mean absolute error (MAE), but also by the variance in the data explained by the model. The latter can be done by using the $R^2$ metric (typically ranged between 0 and 1). A model that always predicts the mean in the data would have an $R^2$ of 0; a model that explain all the variance in the data would have a value of 1. Values below 0 are possible, but point to an unstable model. A random sample of ten percent of the data was used for validation. Results on the validation set are shown in Table \ref{tab:training_results}.

It is evident from Table \ref{tab:training_results} that all models tested struggle to find patterns between the input and the gameplay duration output. On the other hand, even simple models (such as linear regression and the perceptron) can fairly accurately predict the score of player 1. This difference between accuracy of the two outputs is evident from $R_t^2$, as often the model can not explain the variance in the duration data (compared to high $R_s^2$ values). The CNN model can predict both of these outputs more accurately than other models; admittedly, however, the difference with the MLP is small and even simpler models such as the perceptron perform surprisingly well in these tasks. There is a likely explanation for this, as there are significant Pearson correlations between score and 6 class parameters (accuracy, damage, and hit points of both players), and between duration and accuracy of both players. Another factor might be the skewed distribution of game duration in the dataset, which may allow the model to predict values near the mean of the data without a sufficient penalty. This could explain the the simultaneously low error ($MAE_t$) and low explainability ($R_t^2$). Despite the small difference in performance between the CNN and the MLP, the CNN is selected as the model for this work as it allows for image-based feature visualization and attribution, such as activation maximization\cite{erhan2009visualizing} and class activation mapping\cite{selvaraju2016grad}. These processes make it easier to interpret which patterns the model has learned and will be explored in future work.

\section{Generating Character Classes}\label{sec:generating}

With a trained CNN model in place, it is now possible to swiftly test new level and character class combinations. The model acts as a surrogate model of gameplay, and can thus be used in lieu of time-consuming simulations as an evaluation function of e.g. a genetic algorithm. This paper focuses on creating a set of character classes appropriate for a specific level, i.e. changing the character classes' parameters but not the level itself. As an AI-assisted design tool, a human designer can define what ``appropriate'' means in this context. Since the current model outputs both game balance (in terms of the score of player 1) and game duration, the most straightforward approach is to assess a generated pair of character classes based on their distance to a desired game balance and duration. This paper uses the following fitness score to evaluate evolving class matchups, and attempts to minimize it:
\begin{equation}
F(x) = \sqrt{(t(x) - d_{t})^2 + (s(x) - d_{s})^2}
\label{eq:fitness}
\end{equation}
where $d_{t}$ and $d_{s}$ are the desired values for game duration and kill ratio (score) of player 1 respectively, and $t(x)$ and $s(x)$ are the network's predictions of those metrics for a game played by the character classes defined by $x$ on the target map.

Experiments in this paper attempt to generate balanced classes for the levels shown in Fig.~\ref{fig:levels}. The levels in the top row are created by the same level generator used to produce the corpus: they are similar (but not identical) to levels on which the CNN model was trained on. The levels in the bottom row were created by a human designer, and many of them feature a degree of symmetry that is unlikely to be exhibited by the generated levels. Moreover, the human-authored levels attempt to balance the distances from the bases to the powerups and feature explicit level patterns such as arenas, choke points and flanking routes \cite{hullet2010fpspatterns}.

\setlength{\fboxsep}{0px}
\setlength{\fboxrule}{1px}
\newcommand{\mapsize}{0.09}
\begin{figure*}[h!t]
\centering
\setlength{\tabcolsep}{2px}
\begin{tabular}{cc cc cc cc cc} 
\renewcommand*\thesubfigure{\arabic{subfigure}} 
\centering		
\subfloat[G1]{\fbox{\includegraphics[width=\mapsize\textwidth]{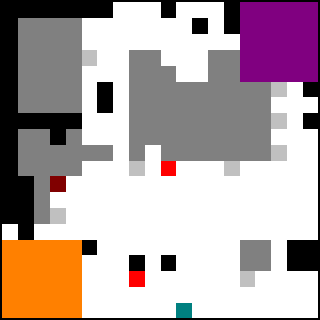}}}&	 	
\subfloat[G2]{\fbox{\includegraphics[width=\mapsize\textwidth]{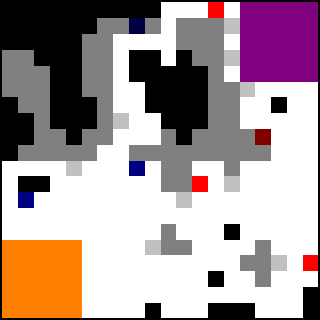}}}&		 	
\subfloat[G3]{\fbox{\includegraphics[width=\mapsize\textwidth]{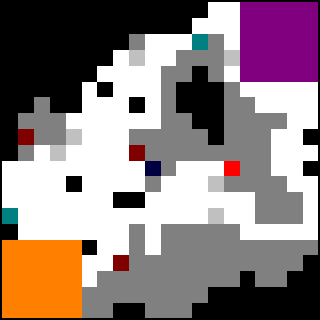}}}& 		
\subfloat[G4]{\fbox{\includegraphics[width=\mapsize\textwidth]{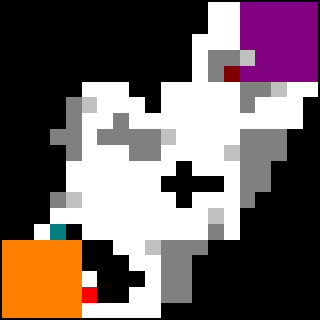}}}&    		
\subfloat[G5]{\fbox{\includegraphics[width=\mapsize\textwidth]{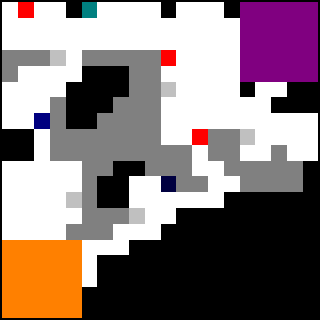}}}& 		
\subfloat[G6]{\fbox{\includegraphics[width=\mapsize\textwidth]{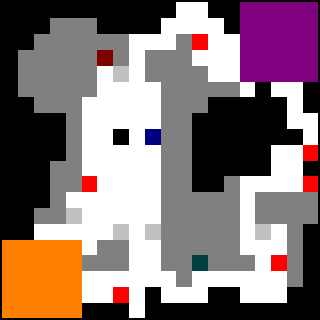}}}&	
\subfloat[G7]{\fbox{\includegraphics[width=\mapsize\textwidth]{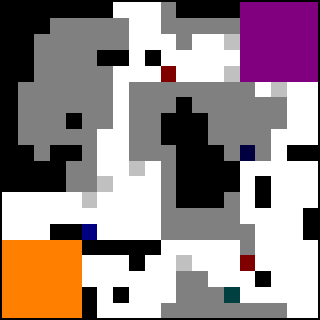}}}&	
\subfloat[G8]{\fbox{\includegraphics[width=\mapsize\textwidth]{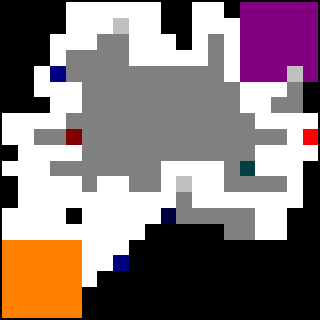}}}&	
\subfloat[G9]{\fbox{\includegraphics[width=\mapsize\textwidth]{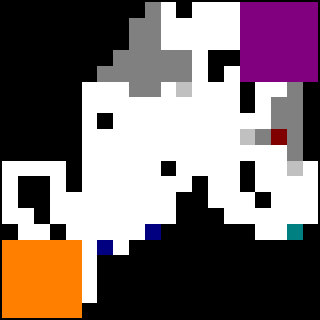}}}&	
\subfloat[G10]{\fbox{\includegraphics[width=\mapsize\textwidth]{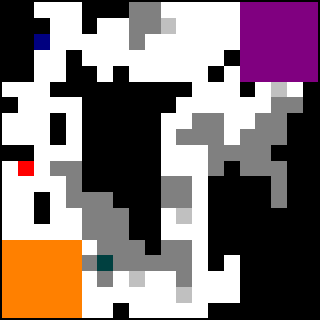}}}	
\\[-5px]
\subfloat[D1]{\fbox{\includegraphics[width=\mapsize\textwidth]{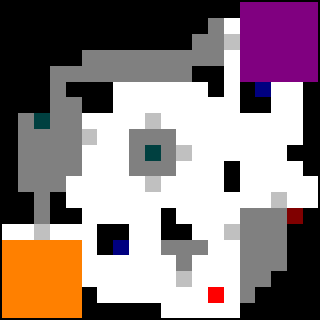}}}& 		
\subfloat[D2]{\fbox{\includegraphics[width=\mapsize\textwidth]{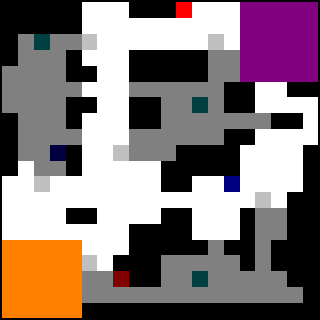}}}& 		
\subfloat[D3]{\fbox{\includegraphics[width=\mapsize\textwidth]{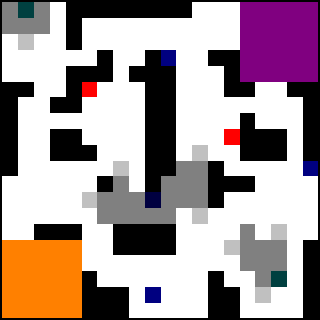}}}&		
\subfloat[D4]{\fbox{\includegraphics[width=\mapsize\textwidth]{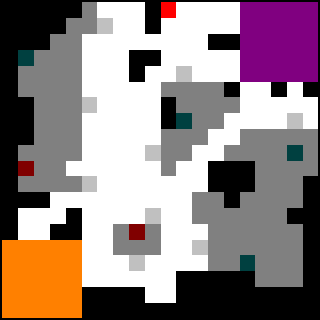}}}& 		
\subfloat[D5]{\fbox{\includegraphics[width=\mapsize\textwidth]{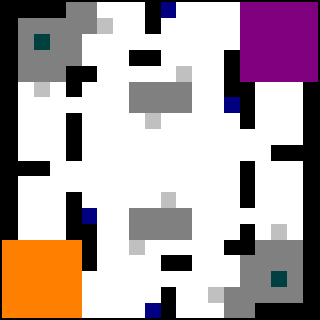}}}& 		
\subfloat[D6]{\fbox{\includegraphics[width=\mapsize\textwidth]{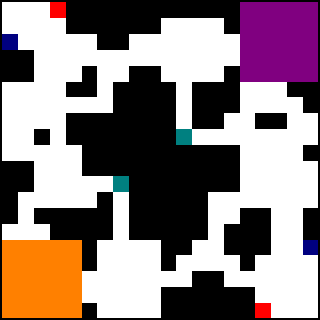}}}& 	
\subfloat[D7]{\fbox{\includegraphics[width=\mapsize\textwidth]{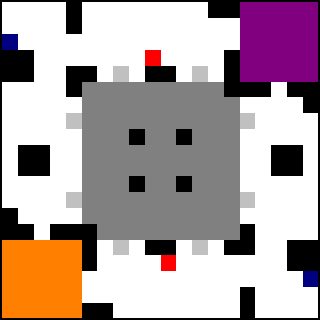}}}&		
\subfloat[D8]{\fbox{\includegraphics[width=\mapsize\textwidth]{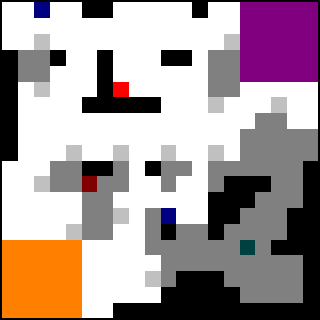}}}&	
\subfloat[D9]{\fbox{\includegraphics[width=\mapsize\textwidth]{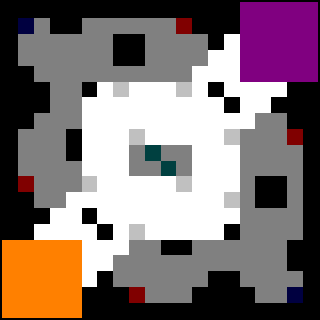}}}&		
\subfloat[D10]{\fbox{\includegraphics[width=\mapsize\textwidth]{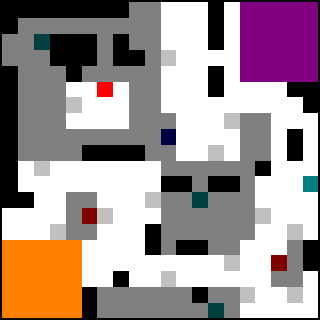}}}		
\\[-5px]
\end{tabular}
\caption{Game levels used for class creation in this paper. Top: Procedurally generated levels. Bottom: Handcrafted levels.}
\label{fig:levels}
\end{figure*}

In each of these levels the goal is to create a balanced matchup, i.e. $d_s=0.5$ in eq.~\eqref{eq:fitness}. In order to test the expressivity of the system, as well as to identify how class parameters are sensitive to different designer priorities, this paper uses three different durations as targets for evolution: \emph{short} (with a target time of 200 sec, i.e. $d_t=0.11$ in the normalized duration), \emph{medium} (target time of 300 sec, i.e. $d_t=0.33$) and \emph{long} (target time of 600 sec, i.e. $d_t=1.00$). Throughout the paper we use the term \emph{short match} to identify a match with classes that were evolved for a short target time (200 sec), regardless of how long the match actually lasted; similarly for medium matches and long matches.

\begin{figure*}[ht]
	\centering
	\subfloat[Generated levels\label{subfig:gt_comparison1}]{\includegraphics[width=0.35\textwidth]{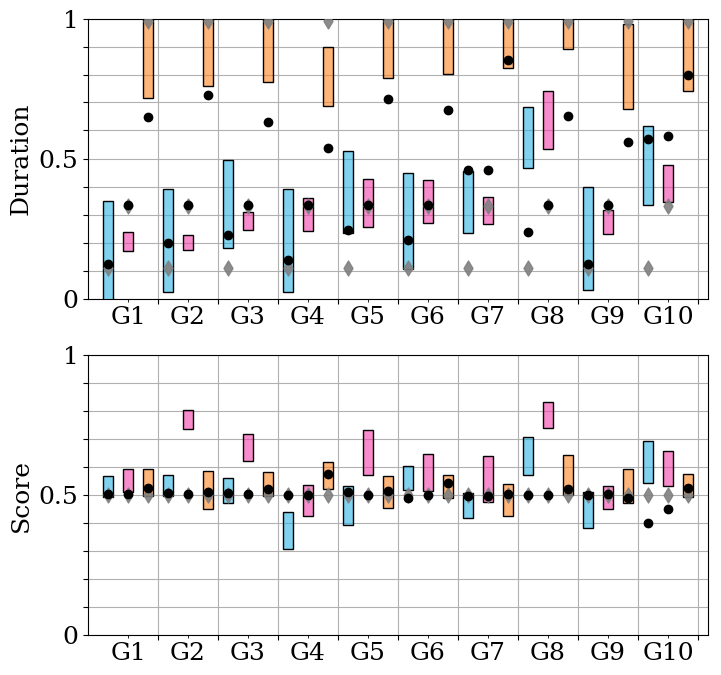}}
	\subfloat[Handcrafted levels\label{subfig:gt_comparison2}]{\includegraphics[width=0.35\textwidth]{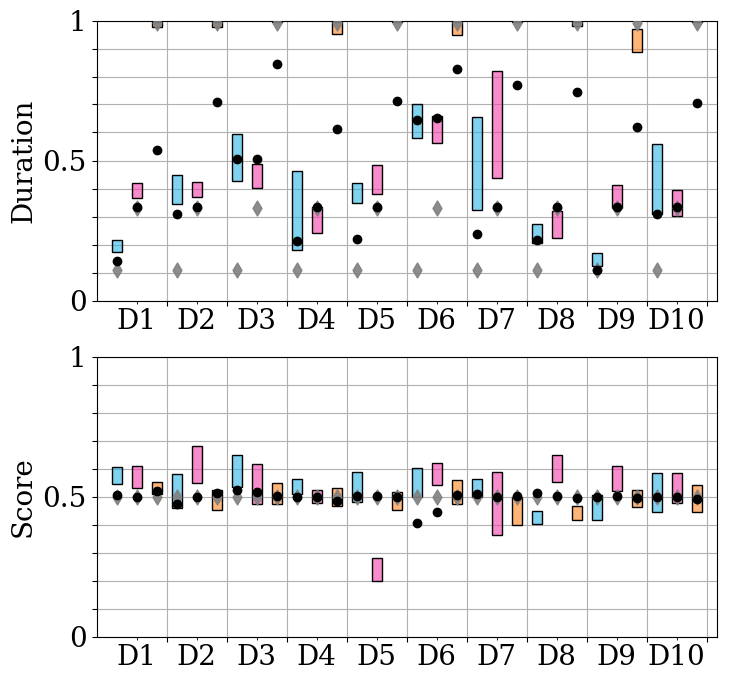}}
\caption{Predicted and actual gameplay parameters for generated class pairs. The actual gameplay parameters are displayed as the upper and lower 95\% confidence bounds from 10 simulated matches per class pair. Class pairs are sorted by level and colored per target duration (blue for short, fuchsia for medium, orange for long). The black dots and gray diamonds respectively indicate the predicted ($p_s$, $p_t$) values and the desired ($d_s$, $d_t$) values.}
\label{fig:gt_comparison}	
\end{figure*}

In order to test a broad set of levels, we ran evolutionary algorithms on each of these 60 testbeds (20 levels of Fig.~\ref{fig:levels}, with 3 desired durations each). Character class pairs are evolved for each provided level, and each desired score ($d_s$) and match duration ($d_t$). In each run, 100 individuals evolve for 100 generations. Initial individuals have random parameters in a genotype of 16 values bound to $[0,1]$ (8 class parameters per player), while individuals are selected for populating the next generation based on roulette wheel selection (to minimize $F$ in eq.~\ref{eq:fitness}). Selected individuals have a 20\% probability of one-point crossover with another selected individual, and can also mutate any of their parameters with a chance of 10\%. Except for weapon range, class parameters are mutated by adding a random number sampled from a normal distribution ($\mu=0,\sigma=0.1$). The mutation of the weapon range randomly changes it into one of the other two possible values (long, medium, short).

Significant findings in experiments assume an accuracy of 95\%.

\subsection{Comparisons with the Ground Truth}

To assess the surrogate-based evolutionary algorithm, we select the best individuals at the end of each evolutionary run (i.e. for one level, $d_t$ and $d_s$ tuple). We test this class pair in game simulations for 10 matches in the same level it was evolved for. The multiple simulations allow us to minimize noise caused by simulation randomness, and to give us a statistical measure of each gameplay parameter through its mean and 95\% confidence interval. Results for the fittest character classes per run, compared to the ground truth (GT) collected from simulated matches, are shown in Fig.~\ref{fig:gt_comparison}. 

It is evident that not all predictions made by the CNN model fall into the confidence bounds of the ground truth from simulations. We consider the model's prediction accurate if its predicted gameplay metric falls within the confidence bounds of the 10 simulations' metrics. Among all 20 levels, estimated duration is accurate in 21 of 60 cases (35\% accuracy) and estimated score is accurate in 36 of 60 cases (60\% accuracy). Gameplay balance is fairly consistently predicted in both generated levels (63\% accuracy) and in designed levels (57\% accuracy). CNN predictions in terms of match duration are less accurate: designed levels are more prone to inaccurate predictions (27\% accuracy) than generated levels (43\% accuracy). Interestingly, in short duration matches the predicted and ground truth durations often match (in 12 of 20 levels); accuracy for match duration drops with medium durations (35\%) and moreso with long durations (10\%). Interestingly, score prediction is less accurate for medium matches (35\%) than for short matches (55\%), but predictions are exceptionally accurate in long matches (90\%).

Besides accuracy of the CNN model when evolving classes, it is important to investigate whether the resulting classes fulfill the broader goals of the designer, in terms of the actual duration and balance of the class matchups. Based on Fig.~\ref{fig:gt_comparison}, it is clear that either predicted or ground truth (GT) values vary depending on the level. Notably, the differences in GT durations are not always perceptible between short and medium matches for the same level. Indeed, only in 2 of the tested maps are the GT durations of short matches significantly different than GT durations for medium matches. 
Across all 20 levels, class matchups for short matches last for 306 sec on  average while for medium matches the average is 319 sec; due to the small difference and high deviations from level to level, the increase in GT duration is not significant between short and medium matches.
On the other hand, the GT duration of long matches is significantly different from GT durations of both short and medium matches in each of the 20 levels tested. 

\begin{table*}
\caption{Differences between mean GT values ($a_t$, $a_s$) and predicted ($p_t$, $p_s$) or desired ($d_t$, $d_s$) values.}
\label{tab:differences}
\begin{tabular}{l|l|c|c|c||c|c|c}
\hline
& & \multicolumn{3}{c||}{Abs. or Eucl. dist. between $(a_t,a_s)$ and $(d_t,d_s)$} & \multicolumn{3}{c}{Abs. or Eucl. dist. between $(a_t,a_s)$ and $(p_t,p_s)$}\\
\hline
Duration & Maps & $|a_t-d_t|$ & $|a_s-d_s|$ & Euclidean & $|a_t-p_t|$ & $|a_s-p_s|$ & Euclidean \\
\hline\hline
Short & Designed & 0.304 $\pm$ 0.232 & 0.052 $\pm$ 0.041 & 0.275 $\pm$ 0.091 & 0.114 $\pm$ 0.110 & 0.068 $\pm$ 0.063 & 0.121 $\pm$ 0.035\\
Short & Generated & 0.243 $\pm$ 0.206 & 0.078 $\pm$ 0.067 & 0.223 $\pm$ 0.082 & 0.136 $\pm$ 0.144 & 0.098 $\pm$ 0.096 & 0.139 $\pm$ 0.060\\
Short & Both & 0.275 $\pm$ 0.185 & 0.066 $\pm$ 0.049 & 0.249 $\pm$ 0.061 & 0.125 $\pm$ 0.109 & 0.084 $\pm$ 0.071 & 0.13 $\pm$ 0.034\\\hline
Medium & Designed & 0.142 $\pm$ 0.143 & 0.105 $\pm$ 0.112 & 0.148 $\pm$ 0.063 & 0.108 $\pm$ 0.128 & 0.110 $\pm$ 0.113 & 0.128 $\pm$ 0.056\\
Medium & Generated & 0.119 $\pm$ 0.133 & 0.149 $\pm$ 0.136 & 0.154 $\pm$ 0.074 & 0.136 $\pm$ 0.132 & 0.154 $\pm$ 0.135 & 0.174 $\pm$ 0.071\\
Medium & Both & 0.131 $\pm$ 0.115 & 0.129 $\pm$ 0.106 & 0.151 $\pm$ 0.047 & 0.123 $\pm$ 0.108 & 0.134 $\pm$ 0.105 & 0.151 $\pm$ 0.045\\\hline
Long & Designed & 0.024 $\pm$ 0.031 & 0.028 $\pm$ 0.028 & 0.031 $\pm$ 0.014 & 0.292 $\pm$ 0.183 & 0.027 $\pm$ 0.027 & 0.281 $\pm$ 0.056\\
Long & Generated & 0.116 $\pm$ 0.090 & 0.040 $\pm$ 0.032 & 0.113 $\pm$ 0.031 & 0.229 $\pm$ 0.136 & 0.024 $\pm$ 0.023 & 0.217 $\pm$ 0.049\\
Long & Both & 0.084 $\pm$ 0.070 & 0.035 $\pm$ 0.026 & 0.072 $\pm$ 0.025 & 0.262 $\pm$ 0.142 & 0.025 $\pm$ 0.021 & 0.249 $\pm$ 0.039\\\hline
Total & Both & 0.182 $\pm$ 0.120 & 0.086 $\pm$ 0.064 & 0.157 $\pm$ 0.032 & 0.182 $\pm$ 0.103 & 0.092 $\pm$ 0.066 & 0.177 $\pm$ 0.026\\
\end{tabular}
\end{table*}

While classes evolved for a long match do indeed fight for longer in simulations, that does not mean that all such simulated matches last the intended 600 seconds. We find that the $d_t$ (i.e. 200 sec, 300 sec, 600 sec) falls within the confidence bounds of GT duration in 25 out of 60 cases; the desired balance (0.5) is within the confidence bounds of the simulations' score ratio in 35 out of 60 cases. The origin of the map (generated or designed) seems to matter little, although designed maps match the $d_t$ only in 8 out of 30 cases, and never accurately match the intended short duration (200 sec). Interestingly, classes evolved for medium matches are the worst at matching $d_s$ (7 of 20 cases). Despite differences from map to map, the classes evolved for medium matches in all 20 maps tend to favor player 1 (with an average score of 0.57, significantly higher than classes evolved for short and long matches).

Another way of evaluating the accuracy of predictions made by the computational model is based on the distance between the mean of the GT values and the predicted values. Similarly, the distance between the mean of the GT values and the desired values is an indicator how close the resulting classes are to what the designer requested. Both of these measures of model accuracy and goal satisfaction are shown in Table \ref{tab:differences}, split along each individual dimension of gameplay (i.e. match duration and time) and combined together as a two-dimensional Euclidean distance. 
A brief overview shows that differences between predicted and GT score values are fairly small, even in the 24 cases where the score prediction falls outside the GT confidence bounds. On the other hand, the difference between predicted duration and GT duration is much more pronounced, especially in long matches. Interestingly, designed maps generally have lower distances than generated matches except in long matches. The large difference between predicted durations and ground truth durations is obvious in Fig.~\ref{fig:gt_comparison}.
The differences between GT values and desired values, however, indicate that in long matches the GT duration is very close to the intended one. This is especially true for designed maps, which is again obvious in Fig.~\ref{fig:gt_comparison}. While the distribution of durations in the training set (see Fig.~\ref{fig:training_data}) does not have many matches with long durations, it is surprisingly easier to produce classes for long matches than it is for short matches as evidenced by the high distance between GT values and desired values (also their Euclidean distance). This contrast between a poorly predicting CNN model in long matches, which however drives evolution towards actually desired durations, is interesting and could be investigated further in future work.

\subsection{Patterns of the Evolved Classes}

Besides the accuracy of the model in terms of ground truth or desired gameplay outcomes, it is important to identify which type of classes are favored for each game level and for each match duration. 

\begin{figure}[t]
	\centering
    \subfloat{
    \includegraphics[width=0.3\textwidth]{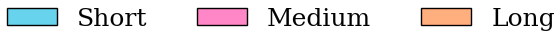}
    }\\
	\subfloat[Class parameters for player 1\label{subfig:rad1}]{%
		\includegraphics[width=0.45\textwidth]{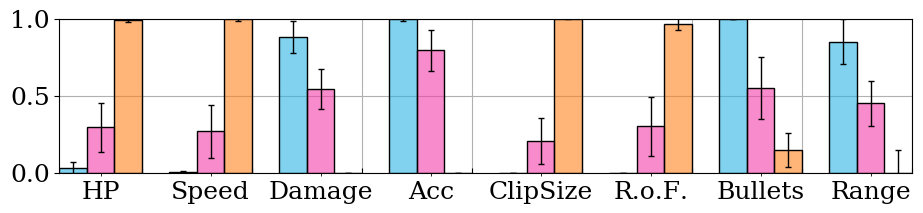}
	}\\
	\subfloat[Class parameters for player 2\label{subfig:rad2}]{%
	\includegraphics[width=0.45\textwidth]{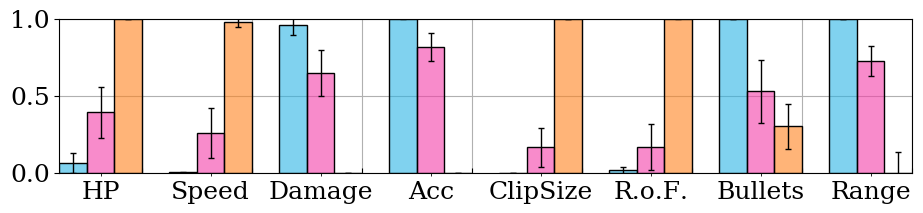}
    }
\caption{The average values for the generated classes' parameters with 95\% confidence interval, colored per duration.}
\label{fig:class_progression}
\end{figure}

Fig.~\ref{fig:class_progression} shows the average parameter values for each player's class across the 20 maps tested. We observe that there are significant differences between parameters in subsequent durations (short to medium, medium to long) in 30 of 32 cases. The patterns are consistent, e.g., hit points keep increasing with longer durations both when moving from short to medium and when moving from medium to long matches. 

Considering the relationship between desired duration ($d_t$) and each player's class parameters, there are significant correlations with 9 of 16 class parameters (indicatively: speed, accuracy, clip size and range of both player 1 and 2). If we consider the mean GT duration of each map tested ($a_t$), there is significant correlation with all 16 parameters. Clearly, patterns learned by the surrogate model drive its choices for class selection. The patterns themselves make sense to a human designer as well: to make gameplay last longer, increasing the hit points of the players and lowering their weapons' damage and accuracy is intuitive. The only surprising trend is an increase in speed for longer matches, which is a consistent finding; we can hypothesize that the high speed coupled with opponents' low accuracy can be used as a secondary defensive mechanism, since a fast-moving player is more difficult to hit.

\subsection{Similarities with Team Fortress 2 classes}

While the quantitative analysis has shown some important trends between the class parameters and durations, it is difficult to estimate in such a way how the generated classes can be played. Instead, we use the parameters of a well-known commercial game to cluster generated classes based on their most similar game class. The game chosen is \emph{Team Fortress 2}, which was an inspiration for much of the work undertaken in this study. Team Fortress 2 (TF2) has 9 classes, each equipped with a signature weapon; these classes have parameters (and scores) comparable to those used in this study. We use 5 of these 9 iconic classes (sniper, soldier, scout, heavy and pyro) as the other classes have special class abilities such as healing or turrets which cannot be captured in our parameter mapping. The only addition made to the data of TF2 classes was accuracy, which was annotated based on the intuition of a TF2 veteran player.

In order to assess the difference between generated and TF2 classes, we use the Euclidean distance between the normalized parameter vector of a player's class and the TF2 class. In this way, we calculate the closest TF2 class (i.e. with smallest distance). However, generated classes with a distance above 1.5 were deemed too different from any TF2 class and are classified as ``undefined''. The threshold was chosen on the one hand to avoid too many generated classes as undefined, while on the other hand to not classify generated classes as TF2 classes based on trivial similarities.

\begin{figure}[t]
	\centering
    \includegraphics[width=\columnwidth]{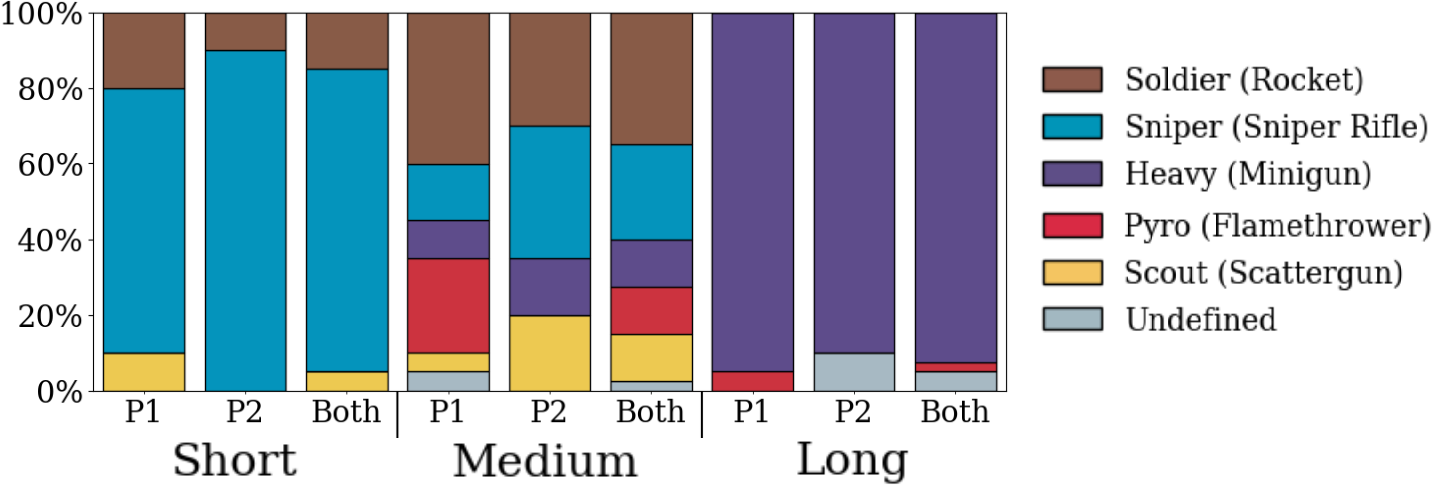}
	\caption{The distribution of classes per player and across players, matched to TF2 classes.}
\label{fig:class_dist}	
\end{figure}

Figure \ref{fig:class_dist} shows the distribution of classes per player and across players. The patterns here are more illustrative than Figure \ref{fig:class_progression}: for short matches, almost all classes evolved for either player are similar to long range classes (sniper, soldier). For long matches, the opposite is true and almost all classes resemble the TF2 ``heavy'' class; this is likely due to the characteristic high HP count and survivability of the heavy class, but also due to the fact that it wields a mini-gun (a highly inaccurate, fast-firing short range weapon) which has similar characteristics with the trends in parameters for long durations in Fig.~\ref{fig:class_dist}. Interestingly, classes evolved for medium durations are the most diverse with a fair distribution between long-range and short-range classes, while 17\% of generated classes for either player are too different to any TF2 class. While the differences between distribution of TF2 classes for player 1 and those for player 2 are not very pronounced, when evolving for medium duration in 65\% of matches the two opponents are mapped to different TF2 classes (compared to 40\% of short matches and 15\% of long matches).

\subsection{Influence of Level Patterns}

Beyond the performance of the genetic algorithm and its accuracy, it is worthwhile to investigate how each level in Fig.~\ref{fig:levels} influences both the accuracy and the ground truth gameplay values of the evolved class pairs. A simple qualitative analysis of Fig.~\ref{fig:gt_comparison} shows that in some levels evolution was unable to find classes for a particularly short duration (based on the ground truth of duration collected from 10 simulations): G8, D6 and D7 particularly stand out as even short matches are around 400 seconds. It is possible that the surrogate model was not able to detect important visual patterns of the level to drive evolution to shorter durations. However, the small difference between predicted and actual durations for D3 or D6, as well as an inspection of the levels themselves in Fig.~\ref{fig:levels} suggests a more likely reason: G6 and D6 are very dense with winding corridors and chokepoints on the ground floor, which lowers the usefulness of long-range weapons while much of the time is spent exploring the level to find an opponent. 

Another interesting dimension is how accuracy (i.e.~whether the prediction is within the 95\% confidence bounds of 10 simulated matches) fluctuates from map to map. Taking into account all 3 intended match durations per map, we observe that predictions in some maps are inaccurate for both duration and score: maps G8 and D1 have no accurate duration predictions and only one accurate score prediction (both in the long match), while D8 has no accurate score predictions and only one accurate duration prediction (in the short match). Other maps as inputs result in inaccurate predictions in one gameplay dimension but are accurate in the other: maps G7, G9 and D10 have accurate score predictions in all 3 evolved class pairs but only one accurate duration prediction (long match for G7, short match for G9, medium match for D10).

\subsection{Comparison with Classes Evolved by MLP}\label{sec:experiment_comparison}
Due to the fairly similar validation accuracies of the best MLP in Table \ref{tab:training_results} with the CNN model, it is worthwhile to investigate whether using an MLP as a surrogate model for evolving character classes would yield similar results. Using the same process for evolving and assessing the final content, the 20 maps of Fig.~\ref{fig:levels} were used as input to evolve 60 character class pairs for balanced short, medium and long matches based on predictions of the MLP. Ground truth values were established on the fittest individuals, and the core findings are compared with the results of the CNN surrogate model.

Classes evolved for the MLP model were accurate (i.e. within the GT confidence bounds) in 36 of 60 instances for predicting score and in 15 of 60 instances for predicting match duration. The score accuracy matches that of the CNN model; duration accuracy drops from 21 (CNN) to 15 (MLP), a relative decrease of 29\%. This is not surprising, since the MLP had a lower $R^2_t$ value than the CNN while both had high $R^2_s$ values (see Table \ref{tab:training_results}). 

\begin{figure}[t]
	\centering
    \includegraphics[width=\columnwidth]{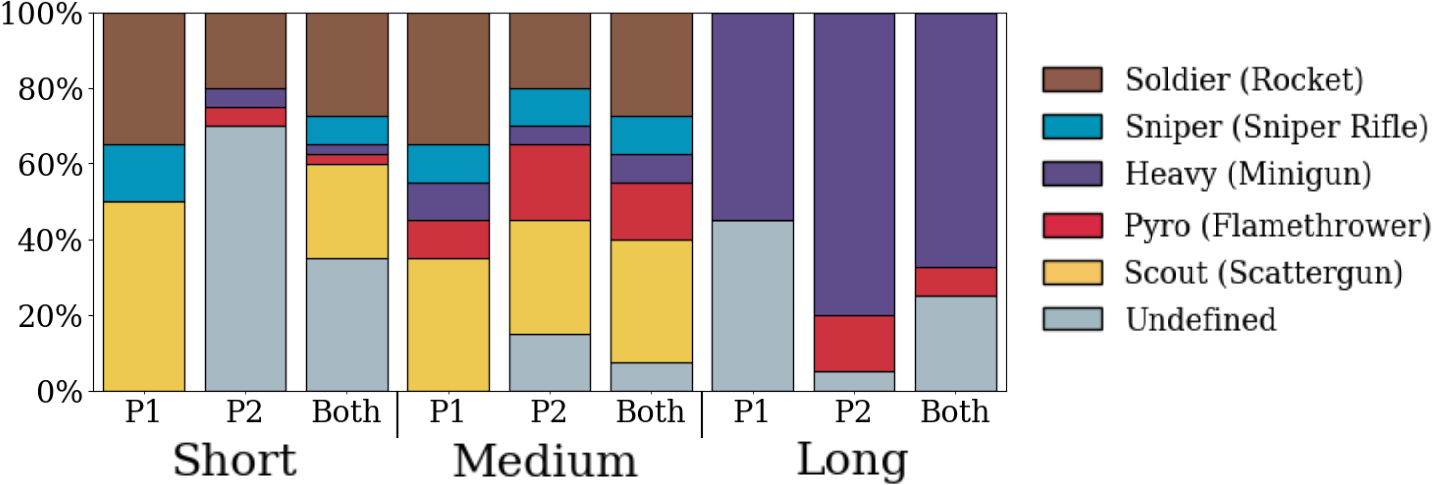}
\caption{The distribution of classes generated by the MLP surrogate model, matched to TF2 classes.}
\label{fig:class_dist_mlp}	
\end{figure}

Classes evolved for the MLP model also have different patterns. Some of the class parameters have diverging trends: e.g. the speed of both players decreases in longer matches (contrary to CNN trends) and rate of fire of both players does not increase in long matches. 
Distribution of MLP-based classes matched to TF2 classes for the same threshold (1.5) are shown in Fig. \ref{fig:class_dist_mlp}. Differences with Fig.~\ref{fig:class_dist} are clear, such as more undefined instances or a limited number of sniper-like classes in short matches.

From this short analysis, it is clear that despite having similar accuracies, the MLP not only underperforms when attempting to predict duration in more cases than the CNN, but it also favors very different patterns in the evolved classes. 

\section{Discussion}\label{sec:discussion}

The research described in this paper has two main goals: to train a computational model which maps game levels, class parameters and gameplay outcomes, and to use this model as a surrogate for simulation-based evaluation of a search-based generator. The deep network described in this paper learned to identify a visual representation of a game level and the parameters of character classes of two players competing in a shooter game deathmatch, and could estimate how the interrelations between level and class characteristics would impact gameplay. The CNN model was shown to be superior ---however slightly--- to simpler and shallower approaches, and its performance as a surrogate model was improved compared to a multi-layer perceptron. As a regression task, however, it was evident when generating classes for specific levels that the predictions were not always matching the gameplay logs collected via simulations. Some level structures were more difficult to parse and offer accurate gameplay predictions than others, mainly when predicting match duration (as expected from the lower $R_t^2$ values during validation). The distribution of durations in the training set may be responsible for this bias, as it is skewed towards values near 300 seconds: the system may often estimate values near the mean duration without suffering from a high error. Future work should attempt to balance the dataset in terms of duration, via oversampling or undersampling. Apart from these inconsistencies when predicting duration, the model was shown almost equally capable of handling handcrafted levels as input ---compared to using generated levels similar to the ones it was trained on. This points to promising applications as a designer tool, as discussed below. 

While the computational designer exploited the learned mappings between three game facets (levels, rules, gameplay) to create character classes for a given level and a given gameplay outcome, an obvious extension of this work would be to generate a level for a given set of classes and given gameplay outcomes. Indeed, \cite{karavolos2018surrogate} showed how existing levels can be evolved to better approach desired gameplay outcomes (e.g.~balance) for a specific class matchup. Future work could generate such levels from scratch rather than adapting existing generated or handcrafted levels. Future work could also generate both classes and levels that suit them for specific gameplay outcomes, orchestrating the generation of multiple facets by keeping in mind their interrelations \cite{bidarra2015orchestration,liapis2014gamecreativity}.

The computational designer's uses are multiple; we have already noted that other facets can also be generated through processes similar to those presented here. The generative potential of the model has been tested in a preliminary fashion in this paper, to create classes from scratch (starting from an initial population of random parameters). Since the model demonstrated that it can handle human-authored levels as input, however, the computational designer could be integrated as a  collaborator to a human designer. This could be done, for instance, by working alongside the designer and starting its evolutionary search from designer-provided starting points, similar to \cite{karavolos2018surrogate,liapis2013sketchbook}, either to create a level or a set of classes. In that role, it would act to refine pre-existing content or ideas of a human author. The CNN model could be used to evaluate human-authored content during the design process, assessing the gameplay outcomes so that the designer could adapt the level or classes towards their own desired outcomes. Moreover, with visualizations popular in deep learning (such as class activation mapping \cite{selvaraju2016grad}), the CNN could highlight to the designer which parameters or level structures increase match duration or a player's advantage. The designer could then use that information to change those parameters or replicate those desirable level patterns.

An obvious shortcoming of the current surrogate model is that it is not always  accurate. While MAE values in Table \ref{tab:training_results} are fairly low, experiments demonstrated that in many cases both score and duration predictions deviated from their simulation-based ground truth. It is difficult to ascertain whether such inaccuracies may have led evolution away from more promising class pairs than the final (tested) ones, i.e. whether it converged to a false optimum provided by the surrogate model \cite{jin2005comprehensive}. An important strength of the surrogate model is its speed when it replaces simulation-based evaluations for a search-based generator \cite{togelius2011searchbased}. Indicatively, one evolutionary run as described in Section \ref{sec:generating} lasts for 50 seconds on a 12-core Intel i7 processing unit; a single death match simulation (optimized to run without graphics, and other speedups) lasts for 50 seconds on the same machine. Assuming that we use the mean gameplay metrics from 10 simulations for each class pairing (as used to calculate the ground truth in Section \ref{sec:generating}), one evolutionary run with 100 individuals evolving for 100 generations would last 58 days; even with a single simulation per individual one run lasts 5.8 days. The overwhelming computational overhead renders testing the surrogate model against a pure simulation-based model unrealistic.

The results of the experiments point to two possible directions for future work. First, discrepancy between prediction and ground truth values may require a more involved re-training process (e.g. balance the dataset towards longer matches, or collect more accurate data by averaging gameplay metrics based on multiple simulations of the same level-class pair). Alternatively, we can re-introduce simulation-based evaluations when the predictive model deems that a map has sufficient quality that a more precise assessment of the gameplay outcomes (via simulations) is warranted. A range of options for when to re-activate simulation-based evaluations exists, such as individual-based, generation-based or adaptive evolution control \cite{jin2005comprehensive}. Second, the evaluation could move away from the Euclidean distance which combines both gameplay outcomes (and treats them as equivalent). A possible side-effect of this aggregation between distance from a desired score and a desired balance is evidenced by the fact that when evolving for long durations, 90\% of score predictions were accurate compared to only 10\% of duration predictions. It is possible that similarity to the intended scores led search away from more promising (and accurate) classes in terms of duration. Instead of aggregating both score and duration predictions into the same fitness, a multi-objective approach \cite{coello2006historical} could treat distance from intended duration and distance from intended score as individual ---possibly conflicting--- objectives. The generator might also leave this trade-off between fidelity to intended duration and fidelity to game balance up to the designer, by presenting the Pareto front (and accompanying classes) to a user for manual selection.

\section{Conclusion}\label{sec:conclusion}
This paper demonstrated how deep learning and evolutionary computation could be exploited to create a computational game designer which can assess game content visually and generate content of one facet based on information of two other facets. Focusing on shooter games, the generator created sets of character classes for two opposing players in a death match; these classes were fitted to a specific game level and intended gameplay outcomes. Future work could exploit this learned model of associations among dissimilar facets to generate content of different types (e.g. levels) or as a co-creator for a human game- or level-designer.

\section*{Acknowledgments}
This research has received funding from the European Union's Horizon 2020 research and innovation programme under grant agreement No 693150.

\bibliographystyle{ACM-Reference-Format}
\bibliography{pairing_full}

\end{document}